\documentclass[runningheads,a4paper]{llncs}

\usepackage{amssymb}
\usepackage{graphicx}
\usepackage{amsmath}
\usepackage{amstext}
\usepackage{subfigure}
\usepackage{enumitem}
\usepackage{url}
\urldef{\mailsa}\path|{alfred.hofmann, ursula.barth, ingrid.haas, frank.holzwarth,|
\urldef{\mailsb}\path|anna.kramer, leonie.kunz, christine.reiss, nicole.sator,|
\urldef{\mailsc}\path|erika.siebert-cole, peter.strasser, lncs}@springer.com|

\usepackage{xspace}

\usepackage{color}

\usepackage{array}
\newcolumntype{x}[1]{>{\centering\arraybackslash\hspace{0pt}}p{#1}}
\usepackage{multirow}
\usepackage{bbding}
\usepackage{graphicx}
\usepackage{subfig}
\usepackage[T1]{fontenc}

\usepackage{graphicx}
\usepackage{caption}

\usepackage{mwe} 
\usepackage{fancyhdr}

\newcommand{\zstroke}{%
  \text{\ooalign{\hidewidth -\kern-.3em-\hidewidth\cr$z$\cr}}%
}

\usepackage{lipsum}

\newcommand*\samethanks[1][\value{footnote}]{\footnotemark[#1]}

\begin{document}

\renewcommand\rightmark{Multi Layered-Parallel Graph Convolutional Network(ML-PGCN) for Disease Prediction}
\mainmatter 
\title{Multi Layered-Parallel Graph Convolutional Network (ML-PGCN) for Disease Prediction}

\titlerunning{Multi Layered-Parallel Graph Convolutional Network (ML-PGCN) for Disease Prediction}

\author{Anees Kazi\inst{1}\thanks{A. Kazi and S. Albarqouni contributed equally to this paper.}\and Shadi Albarqouni\inst{1}\samethanks \and Karsten Kortuem\inst{2}, Nassir Navab\inst{1,3}}
\authorrunning{A. Kazi, S. Albarqouni, K. Kortuem, N. Navab}

\institute{Computer Aided Medical Procedures, Technische Universit\"at M\"unchen, Germany,\\
Augenklinik der Universit\"at, Klinikum der Universit\"at M\"unchen, Germany,\\
Whiting School of Engineering, Johns Hopkins University, USA}

\maketitle

\begin{abstract}
structural data from Electronic Health Records as complementary information to imaging data for disease prediction. We incorporate novel
weighting layer into the Graph Convolutional Networks, which weights
every element of structural data by exploring its relation to underlying
disease. We demonstrate the superiority of our developed technique in
terms of computational speed and obtained encouraging results where
our method outperforms the state-of-the-art methods when applied to
two publicly available datasets ABIDE and Chest X-ray in terms of relative
performance for the accuracy of prediction by 5.31 \% and 8.15 \% and for the area under the ROC curve by 4.96 \% and 10.36 \% respectively. Additionally, the model is lightweight, fast and easily trainable.

\end{abstract}

\section{Introduction}
Structural data(age, gender, weight) from Electronic Health Records (EHRs) are exploited by Computer Aided Systems (CADS) as complementary information for disease prediction. Such systems, however, fail to weight the structural data
based its relevance to the disease at hand. A model capable of evaluating the significance
of every element of the structural data and performing the prediction
task based on the selective and weighted procedure for elements of structural data is required. Such scheme will boost more semantic automatic disease prediction
task 
\par Recently multi-modal data is processed using deep learning methods like  Convolutional Neural Networks(CNNs)\cite{xu2016multimodal}, Autoencoders\cite{ngiam2011multimodal}, Modified Restricted Boltzman Machine\cite{suk2014hierarchical} etc. 
These methods provide richer and discriminant feature space which helps to exploit the global complementary information from available modalities, however, fail to address the problem of unequal relevance.

\par Structural data gives statistical information about the population as a whole. This is taken into consideration more recently using graphs, providing a more semantic way of using multi-modal data\cite{parisot2017spectral,kipf2016semi}. These methods focus more on the association between the subjects with respect to either of the modalities and then solve the tasks such as disease prediction with features from other modalities.

Most recent work by ~\cite{parisot2017spectral} presents an intelligent and novel use case of GCNs for the binary classification task. 
The method proposes to use each structural information to construct an affinity matrix separately and eventually combine them to get the neighborhood graph unlike the conventional methods of fusing the information for the prediction task. This method, however, yields varied results for distinct input neighborhood graph. Each of these affinity graphs and indirectly each element of structural data (age, gender, weight, body-mask ) carries distinct neighborhood relationship and statistical properties with respect to the sample space.

Our motivation is to analyze the impact and relevance of the neighborhood definitions on the final task of disease prediction. In addition to that, we want to investigate whether the relative weighting of meta-data can be done automatically. \textbf{Contributions:} 1)We propose a model capable of incorporating the information of each graph separately, 2)Our design architecture bears a parallel setting of GCN layers 3)We introduce a layer which automatically  learns the weighting of each meta-data with respect to its relevance to the prediction task, 4) Our model outperforms the state of the art method on two but publicly available and challenging databases.\\

\section{Methodology}
\begin{figure}[t]
	\centering 
	\includegraphics[width=0.9\linewidth, height=0.3\textheight]{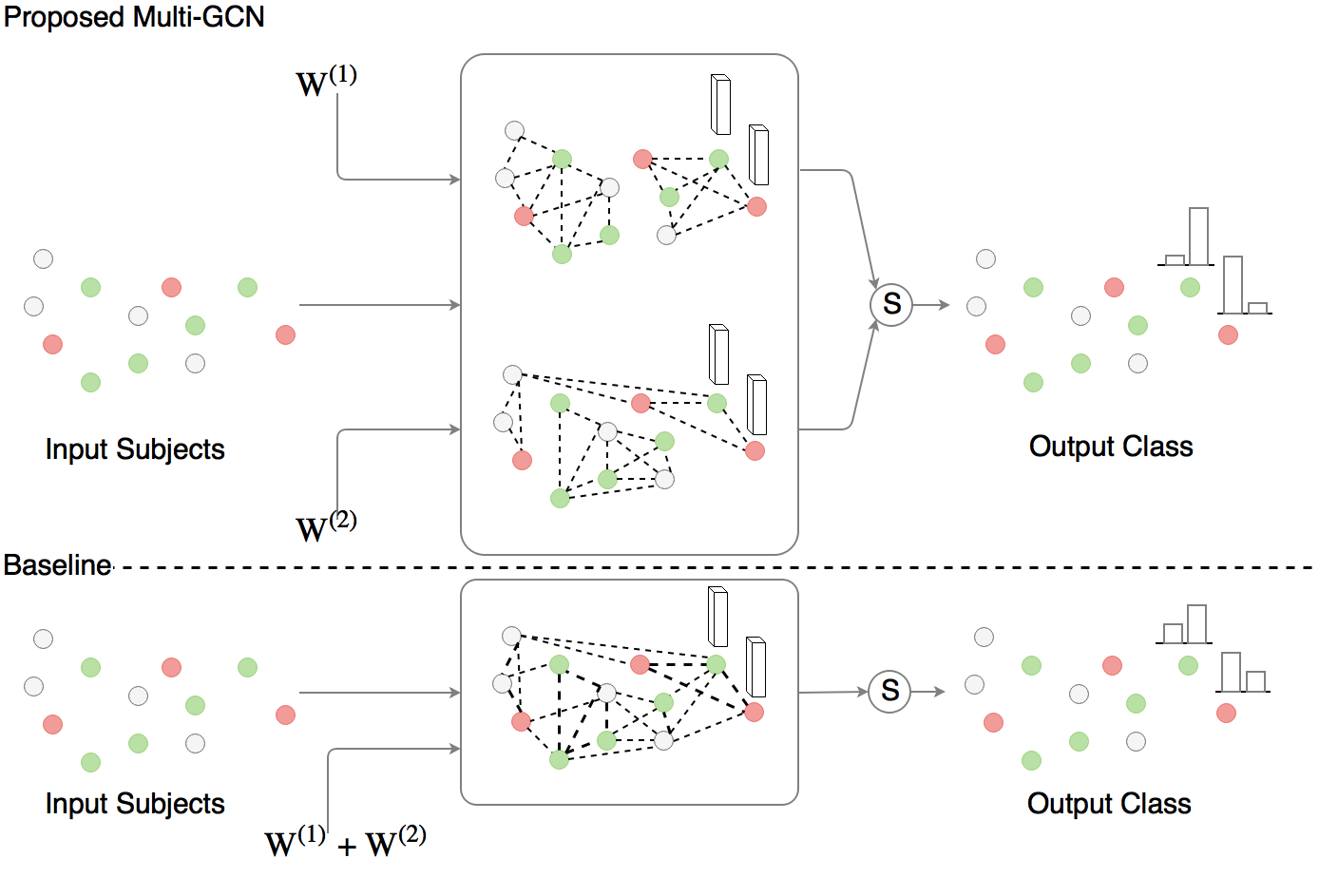}
	\caption{Figure describes the Multi-Layered Parallel Graph Convolutional Network with $M$=2. Two branches have same input features but input affinity matrix.}
	\label{fig:framework}
\end{figure}
\par Considering a database of $N$ subjects, each with data tuple of $d$ dimensional feature vector and $m$ elements from corresponding structural data. Target is to classify the population in $K$ classes with an automatic weighting of structural elements. We adopt such weighting with a parallel setting of Graph Convolutional layers as shown in fig.1. The proposed model incorporates $M$ parallel branches leveraging one of the structural element separately to learn the filters of each branch. Parallel branches consist of 2 Graph Convolutional layers each which are finally fused using the ranking layer. Each branch receives common features for each node. Below we explain the initial setting. Mathematical background on GCN explains the propagation modal and labeling. We provide details on affinity graph construction and finally ranking layer. 

Given a database $\mathcal{D} =\{X, Y\}$ that comprises $N$ subjects $X \in \mathbb{R}^{N\times d}$, each represented by a $d$-dimensional feature vector, and a subset of labeled subjects ($K$-classes) $Y_L \in \mathbb{R}^{L \times K}$, $K$ being one hot vector for our case and $M$ affinity graphs $G^{(m)} \in \mathbb{R}^{N \times N}$ computed from the respective structural element. The objective is to build a model $f(\cdot)$ that learns feature representation that follows the underlying affinity graphs, \emph{i.e}~neighbor subjects should have similar feature representation. This yields a smooth label propagation to the unlabeled set $Y_U \in \mathbb{R}^{U \times K}$ as
\begin{equation}
\hat{Y}= f(X, G^{(m)}; \theta),
\end{equation}
where $\theta$ is the model parameters.
Recent efforts are made to Graph Convolutional Networks (GCN)~\cite{kipf2016semi,defferrard2016convolutional} allowing us to train such a model using 
the layer wise propagation, we adopt the classical GCN propagation model to our ML-PGCN setting as:  
\begin{equation}
H_{l+1}^{(m)} = \sigma \left ( {D^{(m)}}^{-\frac{1}{2}} W^{(m)} {D^{(m)}}^{-\frac{1}{2}} H_{l}^{(m)} \Theta_{l}^{(m)}\right )
\end{equation}
where $W^{(m)}$ is the adjacency matrix of the undirected graph with self connections, $D_{ii}^{(m)} = \sum_{j} W_{ij}^{(m)}$ is the diagonal matrix, $\Theta_{l}^{(m)}$ is the trainable layer-specific weight matrix, which can be derived from a first-order approximation of localized spectral filters on graphs, and $H_{l}^{(m)}$ is the feature representation of previous layer ($H_{0}^{(m)} = X$). It should be noted that ${D^{(m)}}^{-\frac{1}{2}} W^{(m)} {D^{(m)}}^{-\frac{1}{2}}$ is the normalized graph Laplacian, and $\sigma(\cdot)$ is the rectified linear unit (ReLU) function. Using graph spectral theory and polynomial parametrization for $\kappa^{th}$ localized filters\cite{defferrard2016convolutional} the scope of neighborhood is defined. During spectral convolution at any node $n_(i)$, its $\kappa$ neighbors are considered. Readers are referred to \cite{kipf2016semi} for more details on GCN.

\textbf{Affinity Graph Construction:}
Let graph $G^{(m)} = \left \{V, E^{(m)}, W^{(m)} \right \}$ is weighted and undirected consisting of common vertex set $V \in \mathbb{R}^{N}$, and specific edge set $E^{(m)} \in \mathbb{R}^{N \times N}$, and affinity matrix $W^{(m)}\in \mathbb{R}^{N \times N}$. Each reveals distinct intrinsic relationship between the vertices. Edges between vertices are defined based on the given element of structural data information as 

\begin{equation}
E^{(m)}\left( v_i, v_j\right )=  \begin{cases}
1 & if \left | M^{m}\left ( v_i \right ) - M^{m}\left ( v_j \right ) \right |< \beta\\
0 & otherwise
\end{cases}, 
\end{equation}
where $M^{m}(\cdot)$ is the corresponding structural element, \emph{e.g} gender, age, or location, and $\beta$ is a threshold. A similarity metric between the subjects $Sim(v_i,v_j)$, \emph{e.g.} correlation coefficient, is incorporated to weight the edges as 
\begin{equation}
W^{(m)}\left(v_i,v_j\right) =  Sim(v_i,v_j) \circ E^{(m)}\left(v_i,v_j\right),
\end{equation}
where $\circ$ is the Hadamard product.

\textbf{Ranking Layer:}
To rank the structural data elements, we design a linear combination layer that ranks the logits coming from the last hidden layer $H^{(m)} \in \mathbb{R}^{N \times K}$ as
\begin{equation}
	\hat{Y} = \mathbb{S}\left(\sum_{m=1}^{M} \omega_m H^{(m)}\right), 
\end{equation}
where $\omega_m$ is the trainable scalar weight associated with the structural element, $\mathbb{S}(\cdot)$ is the softmax function, and $\hat{Y}$ is the predicted label matrix.\\ 
\textbf{Objective function:}
The model parameters $\theta$ of Graph Convolutional Networks $\Theta^{(m)}$ and Ranking layer $\omega_m$ are updated by back-propagating the gradient of the binary cross entropy loss on labeled data, 
\begin{equation}
\mathcal{L} = -\frac{1}{N}\sum_{n=1}^{L}\sum_{k=1}^{K} y_{nk} \log(\hat{y}_{nk}).
\end{equation}
%
%
%



\section{Experiments and Results}
Our experiments have been designed carefully to firstly investigate the influence of individual affinity graphs on the performance of predictive models. Secondly, to validate our proposed method in the presence of multi-graphs setting compared to the baseline approaches~\cite{parisot2017spectral}. Further, in-depth insight and analysis on ranking structural elements are investigated.\\
\textbf{Dataset:} Our model has been validated on two large and challenging clinical databases for the binary classification task, namely ABIDE~\cite{abraham2017deriving} and Shenzhen Chest X-ray (CXR) Database~\cite{jaeger2014automatic}. Both imaging and non-imaging (structural elements) are provided together with the ground-truth labels. The data is split into training and validation sets (90\%, and 10\%, respectively).\\
\textbf{ABIDE Database\cite{abraham2017deriving}:} 
Autism Brain Imaging Data Exchange (ABIDE) is a grassroots corporation which aggregates data from 20 different sites and openly shares 1112 existing resting-state functional magnetic resonance imaging (R-fMRI) datasets with corresponding structural MRI and structural elements(gender, age) from 539 individuals with ASD and 573 age-matched typical controls.
The database consists of 871 subjects divided into normal (controlled) subjects (468) and abnormal (diseased) ones with ASD (403). It should be noted that for a fair comparison with the baseline, we use the same subset of data and follow the same preprocessing step appeared in~\cite{parisot2017spectral}.\\
\textbf{Shenzhen CXR Database:}
It is created by the National Library of Medicine, Maryland, USA in collaboration with Shenzhen No.3 People's Hospital, Guangdong Medical College, Shenzhen, China. The CXR images are from out-patient clinics and were captured as part of the daily routine using Philips DR Digital Diagnose systems. The database consists of 662 subjects each with structural elements(gender, age) divided into normal (controlled) subjects (326 CXR images) and abnormal (diseased) subjects with Tuberculosis (336 CXR images). All CXR images are passed to a pre-trained AlexNet~\cite{krizhevsky2012imagenet} to extract a lower-dimensional feature representation for each image.\\
\textbf{Implementation:}
All the experiments are implemented in Tensorflow\footnote{\url{www.tesnsorflow.org}} and performed with Nvidia GeForce GTX 1080 Ti 10 GB GPU.
Most of the GCN parameters are adopted from the baseline to obtain the fair comparison.\\
For both datasets we use the network with $m = 2$, each branch having 2 graph convolutional layers each. We keep $d$ = 2000, dropout rate: $0.3$, $\ell_2$- regularisation: $5 \times 10^{− 4}$.  The ABIDE network is trained for 150 epochs, whereas CXR is trained for 500 epochs. We use early stopping criteria to decide the number of epochs for each setting.\\
\textbf{Evaluation:} Our proposed model is evaluated based on the classification accuracy (ACC), and the area under the receiver operating characteristics curve (AUC) on the validation set. Further, we report the paired t-test to measure the statistical significance with a significance level of $5\%$.We perform stratified Monte-Carlo Cross-Validation.\\
\textbf{Notation:} For ABIDE let, $W^{(1)}$ and $W^{(2)}$ be the graph associated with gender and site information respectively. Similarly for CXR let $W^{(1)}$ and $W^{(2)}$ be the graph associated with gender and age.
$\omega^{1} $, $\omega^{2}$  be the corresponding weights.
\begin{figure}[t]
	\centering
	\includegraphics[width=\linewidth]{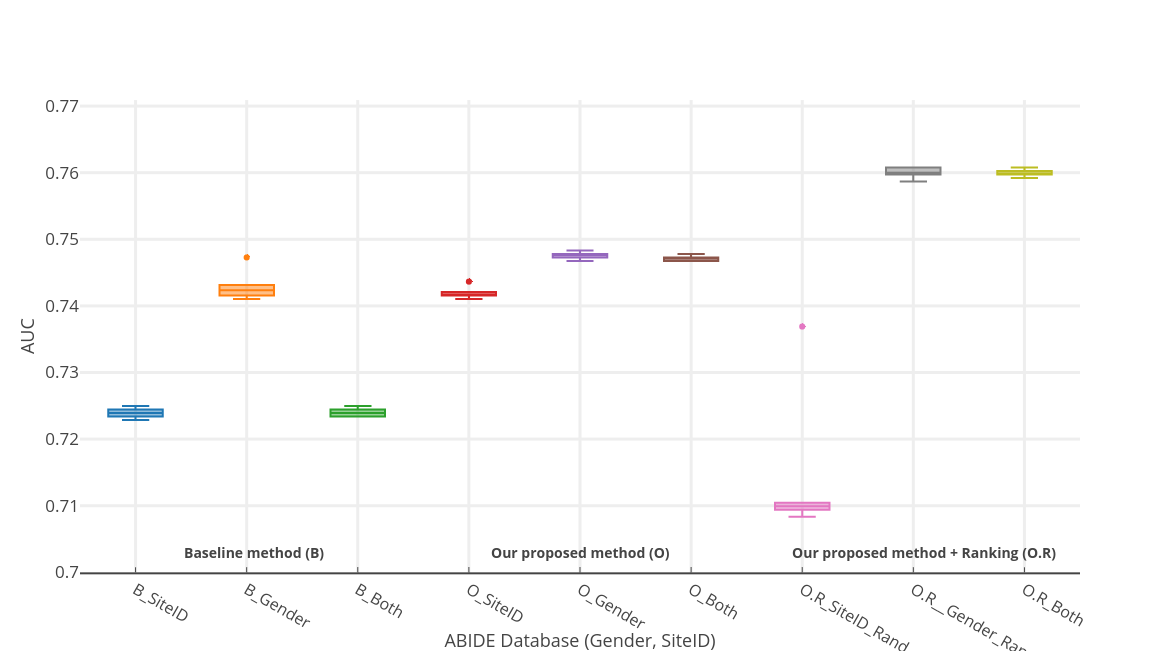}
	\caption{Box plots of AUC reported on ABIDE database for different experiments.}
	\label{fig:ABIDE}
\end{figure}
\begin{figure}[t]
	\centering
	\includegraphics[width=\linewidth]{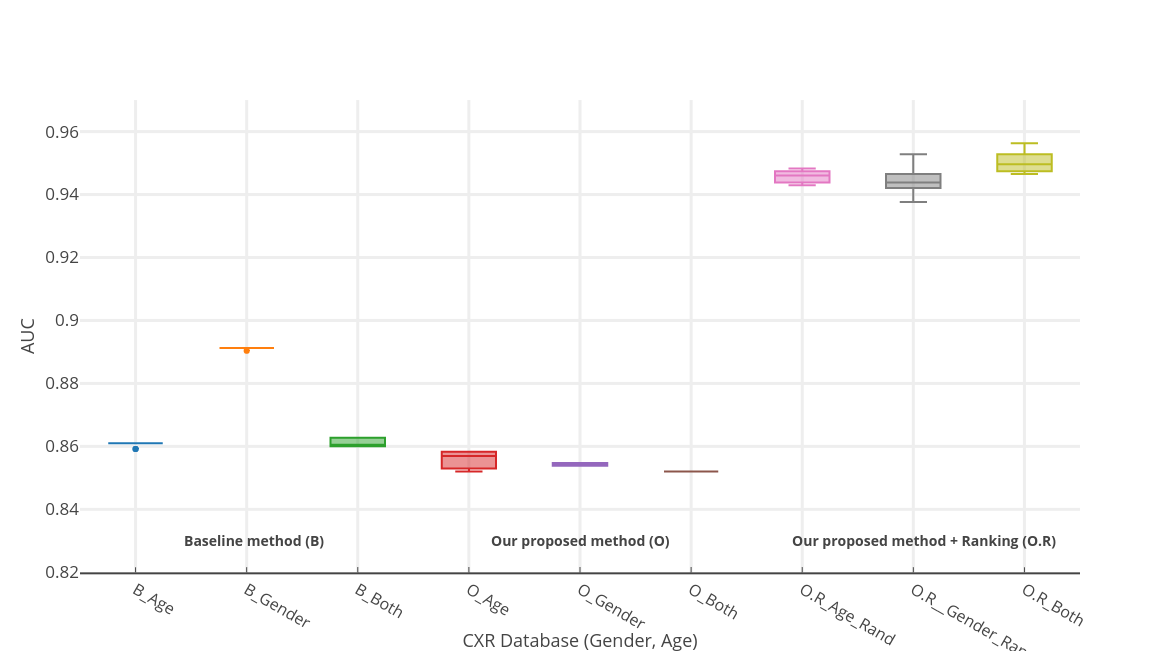}
	\caption{Box plots of AUC reported on CXR database for differest experiments.}
	\label{fig:CXR}
\end{figure}
The accuracies for all the experiments are reported in table~\ref{table:1} for both the datasets.
\\
\textbf{Influence of individual affinity graphs}
We train the conventional baseline method with different input affinity graphs to investigate their individual influence on training and prediction task.
For instance, we investigate if $W^{(1)}$ improves the performance when used alone. The baseline method in Fig.~\ref{fig:ABIDE} and Fig.~\ref{fig:CXR} show the AUCs for ABIDE and CXR datasets, respectively. For ABIDE dataset: \textit{B\_Gender} depicts the best performance($p<0.01$) whereas \textit{B\_Site}  show the worst performance. This shows their individual significance with respect to the prediction task. 
In case of CXR dataset, \textit{B\_Gender} shows the best performance(p<0.01), while \textit{B\_Age} perform same as \textit{B\_Both}. Accuracy and AUC show similar a trend.
For both grid-based convolution and graph-based convolution, the output for each pixel or node depends on features at the node and the neighborhood. Hence the output classification accuracy is a consequence of graph defined. Each graph bearing different neighborhood information performs convolutions with a distinct neighborhood. More semantically relevant neighborhood distinctly separate the clusters based on class.
\\
\textbf{Multi-layer graph model without automatic ranking}
Next, we train proposed ML-PGCN with non-trainable weights for each branch. The motivation of this setting is two-fold. First, to investigate the effect of $W^{(i)}$ for $i = [1,2]$ trained in a parallel setting and explicitly initialize the $[\omega^{1} , \omega^{2} ]$ of the last layer.  ABIDE: 1) $[\omega^{1} , \omega^{2} ]$ = [0, 1] 2) $[\omega^{1} , \omega^{2} ]$ = [1,0]  3) $[\omega^{1} , \omega^{2} ]$ = [0.5,0.5]. We apply similar setting for CXR dataset replacing 'site' with 'age'.
Second, it validates the need of layer which can automatically learn the weighting of two branches. Our proposed method (O) in Fig.~\ref{fig:ABIDE} and Fig.~\ref{fig:CXR} show the AUC for ABIDE and CXR dataset, respectively.
For ABIDE dataset, this setting follows the same trend as the baseline, with  \textit{B\_Gender} performing best($p<$0.01) with both \textit{B\_Site}  and \textit{B\_Both} ), however \textit{B\_Gender} and \textit{B\_Site} perform significantly better with ($p<0.01$)  with respect to \textit{B\_Both}. It is observed that overall our proposed model performs significantly better than baseline method ($p<0.01$). For CXR dataset, our model shows similar performance for all the setting.
\\
\textbf{Multi-layer graph with automatic ranking}
Finally, to validate the importance of ranking the structural elements. The last layer is allowed to train its filters. The motive here is to investigate if the network automatically learns $[\omega^{1} ,\omega^{2} ]$. Our proposed method $+$ Ranking (OR) in Fig.~\ref{fig:ABIDE} and Fig.~\ref{fig:CXR} show the AUC for ABIDE and CXR dataset, respectively.
For ABIDE, our proposed model significantly outperforms all the settings as can be seen in the fig.~\ref{table:1}
($p<0.01$). For CXR dataset, the results from the AUC curve confirms that our model significantly outperforms the baseline model with a huge margin ($p<0.01$)
We also investigate the effect of using the random graph as one of the input. 
The results are demonstrated in Fig.~\ref{fig:ABIDE} and Fig.~\ref{fig:CXR}.
For ABIDE: Results clearly show that \textit{O.R\_Gender\_Rand} outperforms \textit{O.R\_Gender\_siteID} 
For CXR: results with \textit{O.R\_Gender\_Rand} performs better than \textit{O.R\_Gender\_Age}. \textit{O.R\_Gender\_Both} outperforms all the setting as the weighting is automatically done by the ranking layer.
\begin{center}
\begin{table}
\begin{tabular}{ |p{3cm}||p{3cm}|p{3cm}|p{3cm}|  }
	\hline
	\multicolumn{4}{|c|}{ABIDE} \\
	\hline
	Model& Site &Gender&Both\\
	\hline
	Baseline\cite{parisot2017spectral}   & 63.86$\pm$0.012 &\textbf{70.0$\pm$0.008}	&  66.36$\pm$0.005\\
	proposed&   \textbf{68.52$\pm$0.007} & 68.41 $\pm$0.005 &67.50$\pm$0.008\\
	proposed + ranking &- & -& \textbf{69.88$\pm$0.006}
	\\
	\hline
		\multicolumn{4}{|c|}{CXR} \\
	\hline
		Model& Age &Gender&Both\\
	\hline
	Baseline\cite{parisot2017spectral}     &80.59$\pm$0.000
	 & \textbf{82.83$\pm$0.007}& 80.569$\pm$0.000\\
	proposed&   80.59$\pm$0.000 & 80.59$\pm$0.000&80.59$\pm$0.000
	\\
	proposed + ranking& -  & -   &\textbf{87.16$\pm$0.021}
	\\
	\hline
\end{tabular}
\caption{Depicts the mean accuracies from stratified Monte-Carlo Cross Validation for all the setups of experiments }
\label{tab:xyz}
\end{table}
\label{table:1}
\end{center}
\section{Discussion and conclusion:}
\begin{figure}[t]
	\centering 
	\includegraphics[width=0.9\linewidth, height=0.2\textheight]{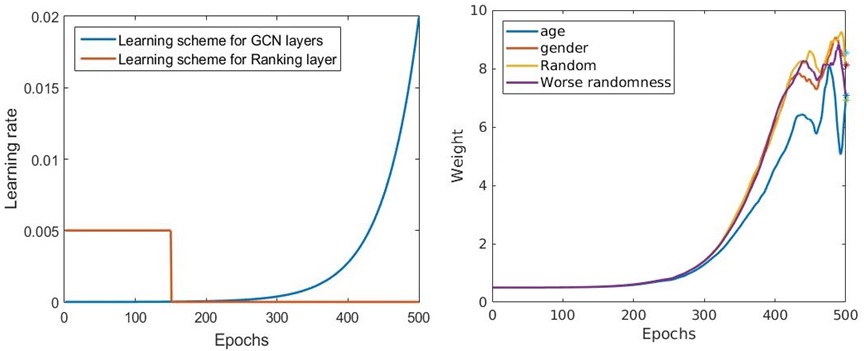}
	\caption{Figure describes the Multi-Layered Parallel Graph Convolutional Network with $M$=2. Two branches have same input features but input affinity matrix.}
	\label{fig:LR_wt}
\end{figure}

All our experiments described, in the previous section, go inline with our hypothesis. GCNs are sensitive to the defined neighborhood. Combination of affinities alters the possible neighborhood between the subjects. Separate spectral convolutions with distinct neighborhood graph in the proposed method without ranking improves the results. Finally, our proposed method with ranking clearly incorporates the unequal contributions of graphs and outperforms all the setups with significant margin. Fig~\ref{fig:LR_wt} depicts weight updates for individual graphs during training which settles down to unequal values based on their contribution. The ratio of training time of our model to baseline is 1.138, making it scalable for a larger number of structural elements.Training GCN layers in the initial epochs followed by training the weighting layer as shown in fig. ~\ref{fig:LR_wt} channelize the learning of weights. Considering our investigation we kept the features at every node simpler, though the definition of such similarity is crucial. Out of sample extension is required for this method, which will boost its usage for the live datasets.



\section{Acknowledgement}
We would like to thank  Freunde und F\"orderer der Augenklinik M\"unchen for the funding.

\bibliography{biblio_new,biblio-macros}
\bibliographystyle{splncs03}

\end{document}